# A CMOS Spiking Neuron for Brain-Inspired Neural Networks with Resistive Synapses and *In-Situ* Learning

Xinyu Wu, *Student Member, IEEE*, Vishal Saxena, *Member, IEEE*, Kehan Zhu, *Student Member, IEEE*, and Sakkarapani Balagopal

*Abstract*— Nano-scale resistive memories are expected to fuel dense integration of electronic synapses for large-scale neuromorphic system. To realize such a brain-inspired computing chip, a compact CMOS spiking neuron that performs *in-situ* learning and computing while driving a large number of resistive synapses is desired. This work presents a novel leaky integrate-and-fire neuron design which implements the dual-mode operation of current integration and synaptic drive, with a single opamp and enables *in-situ* learning with crossbar resistive synapses. The proposed design was implemented in a 0.18μm CMOS technology. Measurements show neuron's ability to drive a thousand resistive synapses, and demonstrate an *in-situ* associative learning. The neuron circuit occupies a small area of 0.01mm² and has an energy-efficiency of 9.3pJ/spike/synapse.

## I. INTRODUCTION

Mimicking one of the most efficient organizational system on the planet, brain-inspired chips can easily accomplish tasks such as object recognition by leveraging their remarkable energy-efficiency and unparalleled pattern matching performance, which is still difficult for modern Von Neumann computers [1], [2]. In a radically different approach than the pervasive von Neumann computers, brain-inspired architectures perform computing tasks by communicating spikes between large network of neurons, which are connected through synapses between each other and locally store memory in form of synaptic strength. To enable such a "brain-chip", neuromorphic architecture combined with CMOS spiking neurons and nano-scale resistive (or memristive) synapses have been proposed [3]–[5]. In these systems, neurons are expected to generate specific spike pulses, drive them into a dense resistive synaptic network, and realize *in-situ* learning with biologically plausible learning rules, e.g. spike-rate and spike-timing dependent plasticity (SRDP and STDP). However, existing silicon neurons either fail to accommodate resistive synapses [6]–[9], or need additional learning circuitry attached to synapses which compromises their potential benefit of high-density integration [10], [11]. An opamp-based neuron appeared in [12] but needs significant circuit overheads to drive a large resistive load which is required by a large-scale system with high synapse integration density [5].

This work presents a CMOS event-driven leaky integrate-and-fire neuron (IFN) circuit that operates in dual-mode for spike integration and large resistive synapse driving when firing, and generates digitally programmable STDP-compatible spikes. Thanks to its elegant reconfigurable topology, the proposed IFN naturally appears as a two-port neural network building block. Furthermore the neuron is able to connect with crossbar array of two-terminal resistive synapses directly and realize biologically plausible *in-situ* learnings, in a similar way as its biological counterpart.

The proposed event-driven neuron has been implemented in a 0.18μm CMOS technology and exhibits capabilities of driving more than one thousand synapses and dense integration with 0.01mm² area. The event-driven neuron design results in an energy-efficiency of 9.3pJ/spike/synapse and provides the flexibility to integrate with a broad range of resistive synapses, which are essentially electronic synapses with conductance-based weight and voltage-based programing thresholds. A test chip containing three of the proposed silicon neurons was fabricated and experimental results demonstrated *in-situ* associative learning with resistive synapses.

The proposed CMOS IFN provides a building block for monolithic integration with nano-scale resistive synapses, to realize a large-scale brain-inspired computing architecture. Such architectures are expected to lead to a new paradigm in energy-efficient and real-time machine learning systems.

## II. BRAIN-INSPIRED NEUROMORPHIC SYSTEM

A basic neuromorphic unit is comprised of several synapses and a neuron block, as shown in Fig. 1. It mimics biological neural cell whereby the synapses receive the synaptic spikes from the other connected neurons and converts them into currents according to their synaptic strength, and the neuron block performs spatio-temporal integration of the spiking pulses and generates output spikes similar to the operation of soma. Further, the dendrites and axon blocks are implemented using interconnect circuits which model the spiking signal

This work is supported in part by the National Science Foundation under the grant CCF-1320987. Xinyu Wu, Vishal Saxena, Kehan Zhu and Sakkarapani Balagopal are with the Electrical and Computer Engineering Department, Boise State University, Boise, ID 83725, USA (e-mail: xinyuwu@u.boisestate.edu, and vishalsaxena@boisestate.edu).

Digital Object Identifier: 10.1109/TCSII.2015.2456372

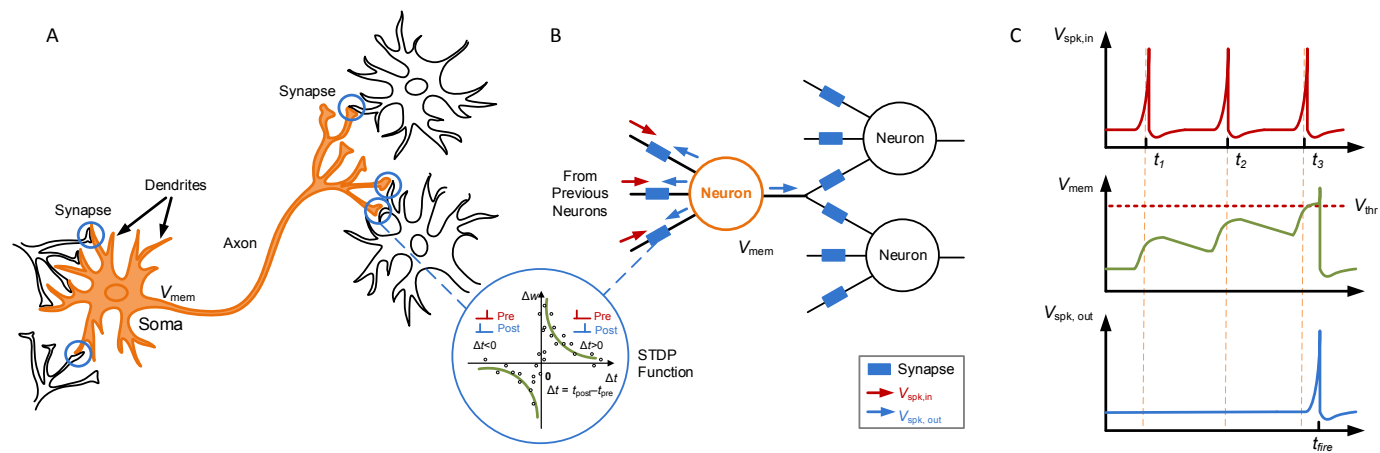

Fig. 1. (A) Simplified diagram of a typical biological neural cell, and (B) a spiking neural cell modeled as a basic neuromorphic unit. Synpatic weight $w$ can be modulated by the pre- and post-synpatic spikes, which is ploted as the STDP function. (C) Working mechnism of a typical integrate and firing neruon. Resistive synapses convert voltage spikes into currents proportional to their synaptic weights. The neuron integrates these current inputs injected into it to change the membrane voltage $V_{mem}$; once $V_{mem}$ crosses a firing threshold $V_{thr}$, the neuron fires and sends a spike $V_{spk,out}$ to pre-synapticand post-synaptic neurons.

propagation through the neuronal fibers.

Similar to a biological synapse, the conductance of a resistive memory (or ReRAM) can be incrementally modified by controlling potential across it. Such resistance modulation has recently been demonstrated in nanoscale two-terminal devices with varied material systems, and the biological plausible STDP and SRDP[1] rules were shown in experiments [3], [5], [13]. STDP states that the synaptic weight is modulated according to the relative timing of the pre- and post-synaptic neuron firing. As illustrated by the embedded picture in Fig. 1, with STDP, repeated pre-synaptic spike arrival before post-synaptic spike leads in a larger synaptic weight; whereas repeated spike arrival after postsynaptic spikes leads to a smaller synaptic weight. The change of the synapse weight $\Delta w$ is generally plotted as a function of the relative timing $\Delta t$ of pre- and post-synaptic spikes and called STDP function.

Ideally, hierarchically connecting a large number of neuron and synapse blocks can realize larger signal processing networks. However, for a long time, the realization of such a large-scale neuromorphic system is difficult due to lack of compact synaptic devices and hardware-friendly network learning method. Until recently, with the advances of understanding synaptic plasticity rules in neuroscience and biological computing communities, hardware *in-situ* synaptic learning becomes feasible. Now, with nanoscale resistive synaptic devices [3] and/or compact CMOS resistive synaptic circuits [14], a versatile CMOS neuron that can interface with these resistive synapses becomes a critical piece to complete the puzzle of a large-scale brain-inspired neuromorphic chip.

## III. CMOS Spiking Neuron Architecture

Since the emergence of neuromorphic engineering, several silicon design styles have appeared in the literature. They model certain aspects of biological neuron, such as sub-threshold biophysically realistic models, compact IFN circuits, switched-capacitor neuron and digital VLSI implementations [4-6]. To allow implementation of a high parallelism system with massive neurons on a single VLSI chip, faithful modeling of biological spiking neurons is prohibitive with limited size and power budgets. Instead, an empirical model as an abstraction for the biological neuron, such as IFN, is employed. However, existing IFN circuits failed to fit into a real large scale neuromorphic system with resistive synapses due to three major challenges: (1) *in-situ* learning in resistive synapses, (2) driving capability and (3) accessory circuits attached to the synapses.

Firstly, conventional IFN circuits are designed to generate spikes to match spiking behaviors of certain biological neurons [6], and then, synapse learning is barely taken into consideration together with neuron circuit. However, brain-inspired learning in resistive synapse requires the neuron to produce spikes with specific shape [5]. Thus, to realize online learning, a pulse generator is needed to produce spikes which are compatible with the electrical properties of the two-terminal resistive synapse. Moreover, a STDP-compatible spike shape with digitally configurable pulse amplitudes and widths is desired to enable the designed silicon neuron to interface with synapse devices with different properties (eg. programing thresholds and operating frequency) and incorporate spike-based learning algorithms, both of which are continuously evolving.

Secondly, in order to integrate currents across several resistive synapses (with $1M\Omega$-$1G\Omega$ resistance range) and drive thousands of these in parallel, the conventional current-input IFN architecture [3] cannot be directly employed; current summing overheads and the large current drive required from the neurons would be prohibitive. Instead, an opamp-based IFN is desirable as it provides the required current summing node as well as a large current drive capability.

Finally, the primary benefit of using nanoscale resistive memory as a synapse is its high integration capability that is quite ideal for resolving the synaptic density challenge in realizing massively parallel neuromorphic systems. For this reason, any accessory circuits attached to synapse for online learning neutralize this benefit and can make resistive synapse less desirable if the accessory circuit occupies large area. Thus, the simplest one wire connection between a synapse to a neuron is desired. To get rid of accessory circuits, current summing and pre-spike driving are needed to be implemented on the same node; same to the post-spike propagation and large current driving. Thus, a compact neuron architecture utilizing opamp driver for both pre- and post-spikes becomes necessary.

---

[1] SRDP is a more generic plasticity rule, and is especially important for short-term plasticity. In this article, we limit our discussion to STDP.

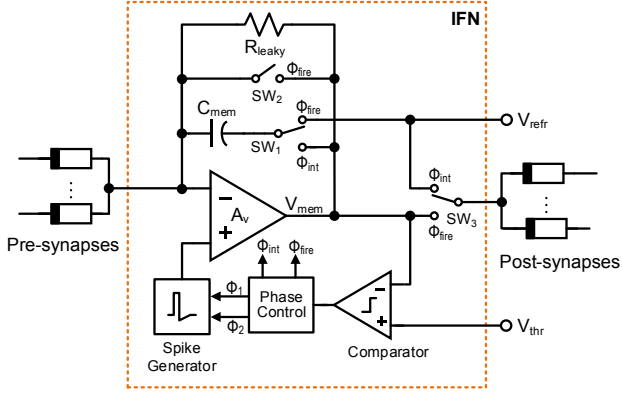

Fig. 2. Block diagram of the proposed event-driven leaky integrate and fire neuron (IFN) circuit.

Fig. 2 shows the circuit schematics of the proposed leaky integrate-and-fire neuron. It is composed of a single-ended opamp, an asynchronous comparator, a phase controller, a spike generator, three analog switches ($SW_1$, $SW_2$ and $SW_3$), a capacitor $C_{mem}$ for integration operation, and a leaky resistor $R_{leaky}$ which is implemented using a MOS transistor in triode. The neuron's dual-mode operation and STDP-compatible spike generating are the key to overcome the three challenges discussed previously.

### A. Event-driven Dual-mode Operation

Event-driven dual-mode operation is realized by using a single opamp that is reconfigured both an integrator, as well as a driver for resistive load during firing events. Here, a power-optimized opamp operates in two asynchronous modes: integration and firing modes, as illustrated in Fig. 3.

In integration mode, phase control signal $\Phi_{int}$ is set to active,

and switch $SW_1$ is set to connect "membrane" capacitor $C_{mem}$ with opamp output. With $\Phi_{fire}$ working as a complementary signal to $\Phi_{int}$, switches $SW_2$ and $SW_3$ are both open. Thanks to the spike generator that is designed to hold a voltage equals to the refractory potential ($V_{refr}$) during the non-firing time, the positive input of opamp is set to voltage $V_{refr}$, which consequently acts as the common mode voltage. With this configuration, the opamp realizes a leaky integrator with the leak-rate controlled by $R_{leaky}$, and charges $C_{mem}$ resulting in a change in the neuron "membrane potential" $V_{mem}$. Next, the neuron sums the currents injected into it, and causes the output voltage moves down. Then, the potential $V_{mem}$ is compared with a threshold $V_{thr}$, crossing which triggers the spike-generation circuit and forces the opamp into the "firing phase."

During the firing-phase, phase signals $\Phi_{fire}$ is set to active and $\Phi_{int}$ is set to inactive which causes switch $SW_2$ is close, and switch $SW_3$ bridges opamp output to post-synapses. Consequently, the opamp is reconfigured as a voltage buffer. STDP spike generator creates the required action potential waveform $V_{spk}$ (discussed later in Section B) and sends to input port of the buffer, which is positive input of the opamp. Noting both pre-synapses and post-synapses are shorted to the buffer output, the neuron propagates post-synaptic spikes in the direction of the input synapses on the same port where currents are summed, and the pre-synaptic spikes in the forward direction on the same node where the post-synapses are driven. At the same time, $SW_1$ is connected to $V_{refr}$, and then discharges the capacitor $C_{mem}$.

For circuit realization, a folded-cascode opamp with a dynamically biased class-AB output stage was employed to accompany the dual-mode operation and optimize energy consumption: the main branch of embedded class-AB stage is shut-off during integration mode using phase signals $\Phi_{int}$ and $\Phi_{fire}$; during firing mode, it is turned-on and provides a rising/falling edge slew rates of 784/500 V/µs to be able to drive a STDP spike with sharp rising and falling edges into resistive loads. The opamp has a DC gain of 72dB and a unity-gain frequency of 274MHz. A dedicated asynchronous comparator is used to compare neuron membrane potential against the firing threshold. To accommodate STDP learning in synapses, comparator hysteresis was traded-off with the speed to support a fast transient response for a pulse-width down to 500ns.

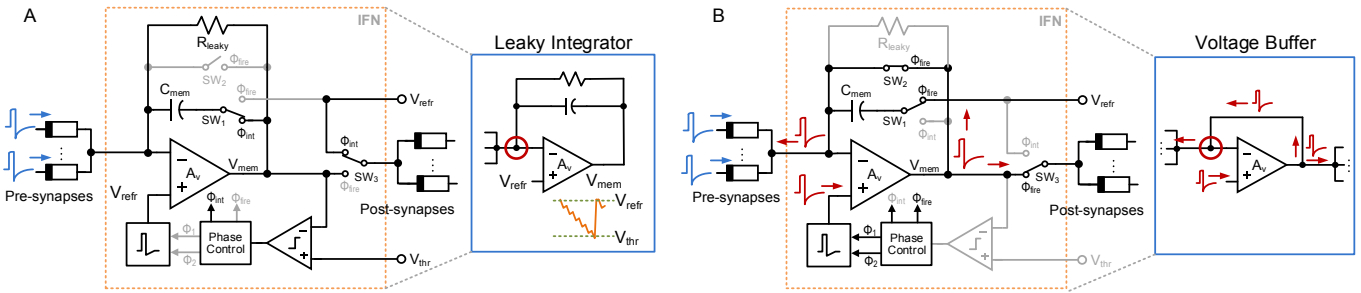

Fig. 3. Dual-mode operation. (A) Integration mode: opamp is configed as a leaky integrator to sum the currents injected into the neuron. Voltages of $V_{refr}$ are held for both pre- and post-resistive synapses. (B) Firing mode: opamp is reconfigured as a voltage buffer to drive resisitive synapses with STDP spikes in both forward and backward directions. Noting backward driving occurs at the same node (circled ) of current summing which enables *in-situ* learning in bare synapses.

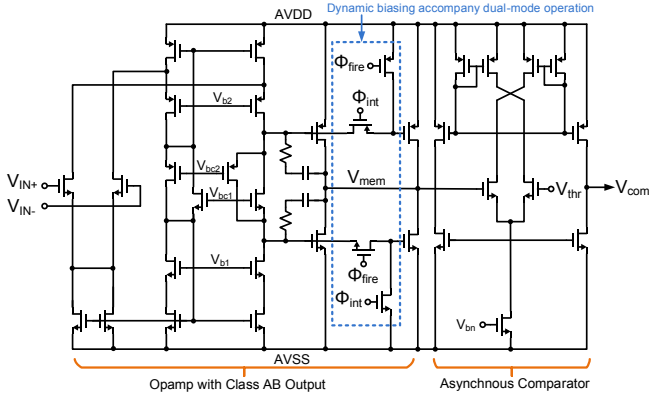

Fig. 4. Circuit details of the embedded opamp and comparator. A dynamically biased class AB stage optimizes power consumption of opamp by reducing the bias current during the integration mode.

### B. STDP-Compatible Spike Generation

The shape of the action potential function $V_{spk}$ strongly influences the resulting STDP function in synapse. A bio-emulative STDP pulse with exponential rising edges is not suitable for circuit implementation. However, a similar STDP learning function in synapse can be achieved with a simpler action potential shape by implementing narrow short positive pulse of large amplitude and a longer relaxing slowly decreasing negative tail [5]. As shown in Fig. 5, the tunable spike generation circuit is designed by selecting between voltage reference levels and an RC discharging waveform for the positive pulse and the negative tail respectively. An on-chip digitally-controlled voltage reference was designed to provide the spike amplitudes $V_a^+$ and $V_a^-$. In addition, digitally configurable capacitor and resistor banks were implemented to provide spike pulse tunability to optimize their transient response to a range of resistive synapse characteristics (e.g., threshold voltage and the program/erase pulse shape as required by the spike-based learning algorithms [3]). Thanks to the dual-mode operation, two connected neurons can drive a pair of these spikes (pre- and post-) into the synapse between them directly. With difference in arriving time ($\Delta t$), pre- and post-synaptic spikes create net potential, $V_{net} = V_{post} - V_{pre}$, across the resistive synapse and modifies the weight if $V_{net}$ over the threshold $V_{t,p}$ or $V_{t,m}$.

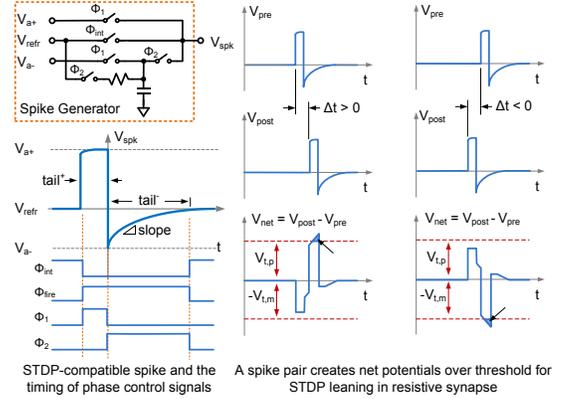

Fig. 5. STDP-compatible spike generation with tunable parameters. These spikes are applied across a resistive synapses and reduce their resistance if $V_{net} > V_{t,p}$, or increase their resistance if $V_{net} < -V_{t,m}$.

For controlling the spike generation, a compact digital phase control circuit generates two non-overlapping control signals, $\Phi_{int}$ and $\Phi_{fire}$, which switch the IFN between the two operation modes. Further, together with another two signals implemented using one-shot pulse circuits, $\Phi_1$ for positive tail and $\Phi_2$ for negative tail, the spike generation timing is precisely defined as seen in Fig. 5.

## IV. MEASUREMENT RESULTS

A test chip was fabricated in a 0.18μm CMOS process, and its micrograph is shown in Fig. 6. The active area of the chip includes circuitries of three neurons that each occupies 0.01mm², digital configurable capacitor and resistor banks, biasing and voltage reference circuitries, and a digital interface. The test-chip also includes three 8×8 on-chip tungsten electrode arrays, the option of resistive synapses integration to be bonded externally and/or fabricated on the CMOS chip using a back-end-of-the-line (BEOL) process.

Fig. 7A shows the measured typical output spike when driving a load with resistance of 1kΩ in parallel with capacitance of 20pF, which is equivalent to few thousands of resistive synapses with 1MΩ nominal. Fig. 7B illustrates such a spike pair that was applied across a synapse with $\Delta t$ around 0.5µs. In a 0.4µs time-window, the spikes created a net potential greater than the synaptic modification threshold, $V_{\text{t,p}} = 340$mV. For smaller loading, the spike has sharper rise and fall edges which cause a greater peak net potential in STDP pair; whereas, for even larger loading, slower rise and fall edges could lead to an under-threshold net potential in STDP pair. The power consumed in the neuron for synaptic plasticity is 9.3pJ/spike/synapse (in terms of 1,000 synapses each having around 1MΩ resistance). Thanks to the on-chip digital tunability, the design interfaces with a broad range of resistive synapses and device material systems. The tunable chip also serves as a platform for characterizing STDP response of resistive memories with several material systems. The design also interfaces with a compact CMOS emulator circuit based on a memory controlled varistor implementation of resistive synapses with STDP learning ability [13].

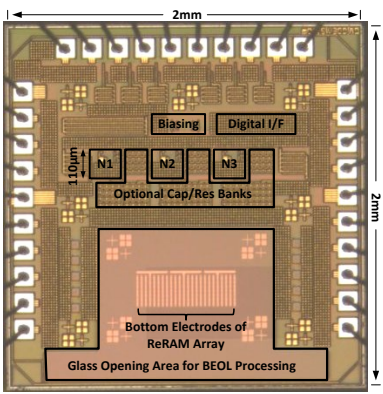

Fig. 6. Micrograph of the test chip in 180nm CMOS. N1, N2 and N3 are three silicon neurons. Biasing is biasing and voltage reference circuitries, and Digital I/F is the digital interface.

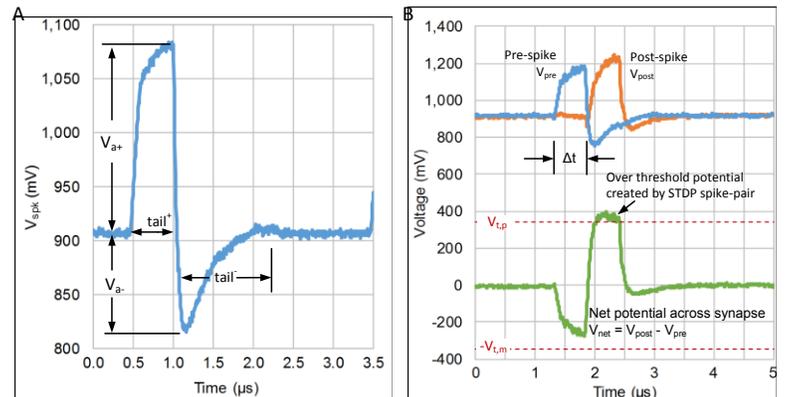

Fig. 7. Measured (A) typical neuron output spike, and (B) the over-threshold net potential across resistive synapse created by a STDP spike pair from pre- and post-synaptic neurons (N3 and N1 on the fabricated chip).

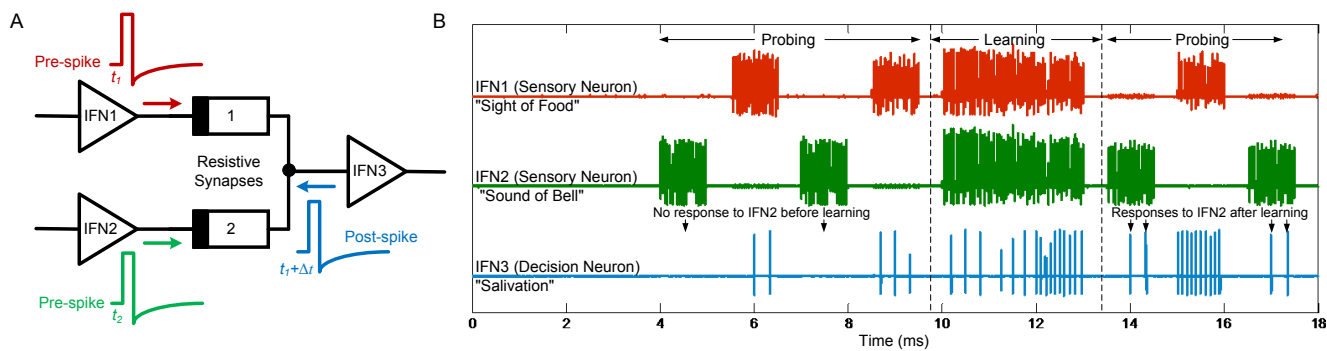

Fig. 8. (A) A spiking network with three neurons and two synapses can emulate the simplest associative leaning of a Pavlov's dog, and (B) an experimental demonstration of associative leaning using the fabricated chip. By simultaneously applying stimulations to both IFN1 and IFN2, synapse 2 was strengthened with STDP learning, which carried larger currents with spike and caused IFN3 responded to IFN2 inputs independently after learning.

Finally, an associative learning was experimentally demonstrated by a neural network with two input neurons for sensory and one output neurons for association decision, as shown in Fig. 8A, also known as the Pavlov's dog [2]. Associative learning is especially important as it is believed to be behind how brains correlate individual events and how neural networks perform certain tasks very effectively. With external resistive synapse emulator ($V_{\text{t,p}} = 340$mV), synapse 1 was initialed to 51kΩ, and synapse 2 was initialed to 1MΩ), associative learning with the fabricated chip was measured and plotted. As shown in Fig. 8B, before learning, the "salivation" neuron (IFN3) only responded to the input from the "sight of food" neuron (IFN1). By simultaneously applying stimulations to both "sight of food" neuron and "sound" neuron (IFN2), synapse between IFN2 and IFN3 was strengthened (lower resistance) with STDP. Then stimulus from the "sound of bell" neuron alone was able to excite the "salivation" neuron, therefore establishing an association between the conditioned and unconditioned stimuli.

## V. Discussion

It is worth noting that the current nano-scale resistive devices suffer from several challenges: device variability, sneak-paths and crosstalk. These effects can be detrimental network-level learning. Furthermore, conventional issues in scaled VLSI systems, e.g. interconnects, power distribution and signal integrity, must be considered when implementing a large-scale brain-inspired neuromorphic chip. As the fabricated chip was designed with the primary purpose to verify the proposed neuron architecture with a range of devices, thus there is further scope for optimizing the layout size and power.

## VI. Conclusions

A compact spiking leaky integrate-and-fire neuron circuit capable of driving a large number of bare resistive synapses and realizing *in-situ* STDP learning is proposed. Measurement results with a fabricated test chip in a standard 0.18µm CMOS process verified its functionality. Further, an associative learning experiment demonstrated the *in-situ* learning without any additional training circuitry. Thanks to its unique topology and dual-mode operation, the proposed CMOS neuron contributes a critical building block to integrate dense resistive synapses for large-scale brain-inspired neuromorphic systems.